\documentclass{article}
\usepackage{spconf,amsmath,graphicx}

\usepackage{epsfig}

\usepackage{array} 
\usepackage{makecell,multirow,diagbox}

\usepackage{caption}
\usepackage{cite}
\usepackage{amssymb}
\usepackage{bm}
\usepackage{color}
\usepackage{subfigure}

\usepackage{float}
\usepackage{booktabs}

\usepackage{epstopdf}

\usepackage[pagebackref=true,breaklinks=true,letterpaper=true,colorlinks,bookmarks=false,citecolor=blue,linkcolor=blue]{hyperref}

\usepackage{CJK}
\usepackage[top=2cm, bottom=2cm, left=2cm, right=2cm]{geometry}
\usepackage{algorithm}
\usepackage{algorithmicx}
\usepackage{algpseudocode}
\usepackage{amsmath}

\floatname{algorithm}{Algorithm}
\renewcommand{\algorithmicrequire}{\textbf{Input:}}

\title{DEEP NEURAL NETWORK BASED SPARSE MEASUREMENT MATRIX FOR IMAGE COMPRESSED SENSING}
\name{Wenxue Cui$^1$, Feng Jiang$^1$, Xinwei Gao$^{1,2}$, Wen Tao$^1$, Debin Zhao$^1$}
\address{1.Department of Computer Science and Technology, Harbin Institute of Technology, Harbin, China \\
2.Wechat Business Group, Tencent, Shenzhen, China }
\begin{document}
%
\maketitle
\begin{abstract}
Gaussian random matrix (GRM) has been widely used to generate linear measurements in compressed sensing (CS) of natural images. However, there actually exist two disadvantages with GRM in practice. One is that GRM has large memory requirement and high computational complexity, which restrict the applications of CS. Another is that the CS measurements randomly obtained by GRM cannot provide sufficient reconstruction performances. In this paper, a Deep neural network based Sparse Measurement Matrix (DSMM) is learned by the proposed convolutional network to reduce the sampling computational complexity and improve the CS reconstruction performance. Two sub-networks are included in the proposed network, which are the sampling sub-network and the reconstruction sub-network. In the sampling sub-network, the sparsity and the normalization are both considered by the limitation of the storage and the computational complexity. In order to improve the CS reconstruction performance, a reconstruction sub-network are introduced to help enhance the sampling sub-network. So by the offline iterative training of the proposed end-to-end network, the DSMM is generated for accurate measurement and excellent reconstruction. Experimental results demonstrate that the proposed DSMM outperforms GRM greatly on representative CS reconstruction methods

\end{abstract}

\begin{figure*}[!t]
\centering
\includegraphics[width=\textwidth]{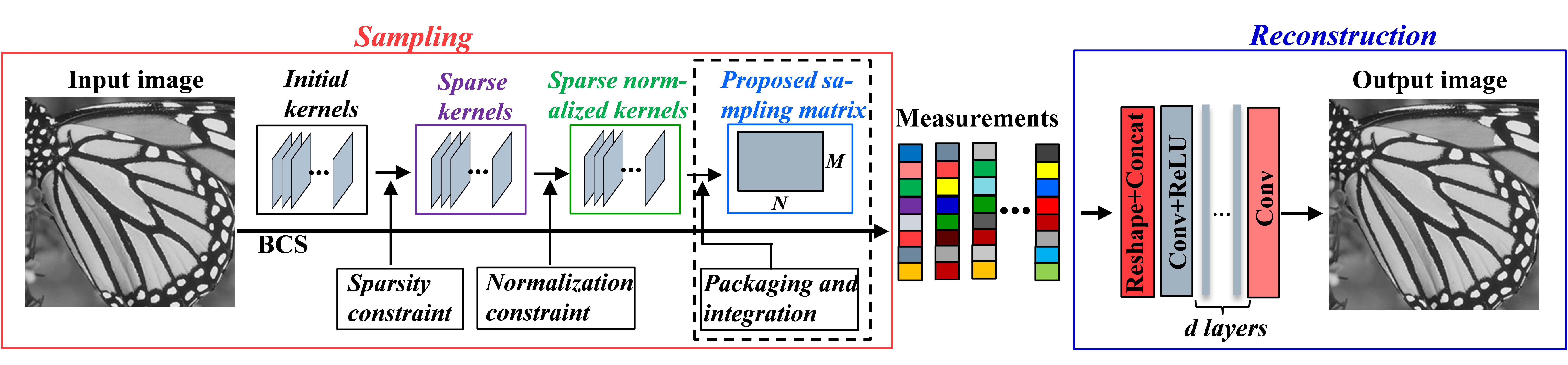}
\vskip -0.1cm \caption{The framework of proposed deep neural network for generating DSMM. The components in the dashed are executed only for the testing phase.}
\label{Fig:fig1}
\end{figure*}

\section{Introduction}
\label{sec:intro}
The Compressed Sensing (CS) theory~\cite{candes2006robust, donoho2006compressed} demonstrates that if a signal is sparse in a certain domain $\boldsymbol{\Psi$}, it can be recovered with high probability from a small number of random linear measurements less than that of Nyquist sampling theorem. Mathematically, the measurements are obtained by the following linear transformation
\begin{equation}
\label{equ1}
  \boldsymbol{y} = \boldsymbol{\Phi} \boldsymbol{x}+\boldsymbol{e}
\end{equation}
where $\boldsymbol{x}\in \mathbb{R}^{N}$ is lexicographically stacked representations of the original image and $\boldsymbol{y}\in \mathbb{R}^{M}$ is the CS measurements observed by a $M\times N$ measurement matrix $\boldsymbol{\Phi$}, $(M\ll N)$. $\boldsymbol{e}\in \mathbb{R}^{M}$ indicates noise. CS aims to recover the signal $\boldsymbol{x}$ from its measurements $\boldsymbol{y}$, which usually consists of the sampling stage and the reconstruction stage. In the study of CS, an excellent measurement matrix in terms of memory requirement and reconstruction performance is the first prerequisite of the reconstruction. How to design a good measurement matrix $\boldsymbol{\Phi$} in terms of both the storage space and the reconstruction accuracy is still a challenge. In the past few years, a large number of efforts have been devoted to it.

In most CS literatures~\cite{li1tval3,chen2011compressed,zhang2012compressed,zhang2014group}, Gaussian random matrix (GRM) is utilized in sampling stage, which requires a vast storage and high computational complexity in most cases. Some structural sampling matrices are proposed in~\cite{dinh2013measurement,gao2015block}, which is based on Block Compressed Sensing (BCS)~\cite{gan2007block}. Recently, several Deep Neural Network (DNN) based methods~\cite{adler2016deep,shi2017deep} are proposed to learn more accurate sampling matrices. However, these DNN-based sampling matrices have two main
disadvantages. One is that these sampling matrices are not normalized while normalization is needed to limit the range of measurements in most algorithms. The other one is these DNN-based methods pay more attention to the accuracy, while ignoring the large storage overhead and expensive computing costs, which leads to the slowness of sampling speed. Getting a superior sampling matrix is still a challenging problem.

For CS reconstruction, many algorithms~\cite{li1tval3,chen2011compressed,zhang2012compressed,zhang2014group} have been developed in the past decade. Most of these methods exploit some structured sparsities as image priors. For example, $\boldsymbol{x}$ is sparse in the domain $\boldsymbol{\Psi}$. Therefore, the CS reconstruction can be implemented by solving a sparsity-regularized optimization problem
\begin{eqnarray}
\label{equ2}
   \tilde{\boldsymbol{x}} = \mathop{\arg\min}_{\mathop{\boldsymbol{x}}}{\frac{1}{2}\|\mathop{\boldsymbol{\Phi}}\mathop{\boldsymbol{x}} - \mathop{\boldsymbol{y}}\|_2^{2}+\lambda\|\mathop{\boldsymbol{\Psi}}\mathop{\boldsymbol{x}}\|_{l_{p}}}
\end{eqnarray}
where $\lambda$ is the regularization parameter to control the tradeoff of fidelity term and sparsity term. These methods all utilize the gaussian random matrix (GRM) as their sampling matrix, which limits their reconstruction ability enormously.

To overcome the shortcomings of the aforementioned sampling matrices for CS, we propose a Deep neural network based Sparse Measurement Matrix (DSMM), which can be learned from a convolution network. In this network, a sampling sub-network and a reconstruction sub-network are designed, which harmonize with each other by an end-to-end training metric. In the sampling sub-network, a novel convolutional layer with sparsity constraint and normalization constraint on parameters is proposed to learn the target sampling matrix. In the reconstruction sub-network, a ``reshape+concat'' layer~\cite{shi2017deep} and several convolutional layers are utilized to reconstruct original images from the measurements, which are obtained from the sampling sub-network. The proposed framework is shown in Fig.~\ref{Fig:fig1}. In the training process, these two sub-networks are stimulated by each other. Specifically, on one hand, in the reconstruction sub-network, images are recovered with the measurements, which are obtained by the sampling sub-network. On the other hand, the sampling sub-network is guided by the reconstruction sub-network for more accurate sampling matrix. By jointly utilizing two sub-networks, the entire network can be trained in the form of end-to-end metric with a joint loss function.

The contributions of this paper are summarized below:
\begin{itemize}
\item The proposed DSMM is sparse, which significantly reduces both the memory requirement and the computation complexity.
\item In the training process, a reconstruction sub-network is introduced to guide the optimization of sampling sub-network and a stable reconstruction performance is acquired by using the learned sampling matrix on representative CS reconstruction methods.
\item Experimental results demonstrate that the proposed DSMM provides a significant quality improvement compared against GRM.
\end{itemize}

\section{Proposed Method}
\label{sec:format}

In this section, we describe the methodology of the proposed method including the sampling sub-network and the reconstruction sub-network.

\begin{figure}[b]
\vskip -0.3cm
\begin{minipage}[t]{0.235\textwidth}
\centering
\includegraphics[width=1.46in]{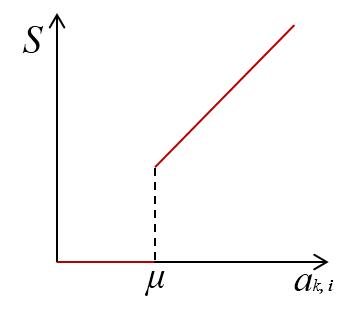}

\end{minipage}
\begin{minipage}[t]{0.235\textwidth}
\centering
\includegraphics[width=1.46in]{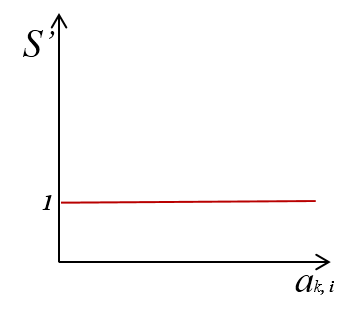}
\end{minipage}
\vskip -0.4cm \caption{Sparsity constraint operation and its approximate derivative}
\label{Fig:fig2}
\end{figure}

\subsection{Sampling Sub-network}
In traditional block-based compressed sensing (BCS)~\cite{gan2007block}, each row of the sampling matrix $\boldsymbol{\Phi}$ can be considered as a filter. Therefore, the sampling process can be mimicked using a convolutional layer~\cite{adler2016deep,shi2017deep}. Specifically, given an image $I$ with size $w\times h$, the measurement vectors of $I$ can be obtained by applying a convolutional operator to the image $\mathcal{Y}=\mathcal{C}_{B}(I)$ where $\mathcal{Y}$ is the set of measurement vectors and $\mathcal{C}_{B}(\cdot)$ denotes the convolution operation with kernel size $B\times B$ and stride size $B\times B$. The block size is $N_{B} = B\times B$ and the dimension of the measurements for each block is $N_{b} = \lfloor\frac{M}{N}N_{B}\rfloor$. However, these DNN-based methods pay more attention to the reconstruction accuracy, while ignoring some essential properties and several major drawbacks for the measurement matrix. In this paper, a sparse normalized measurement matrix is produced by using a novel convolutional layer with a sparsity constraint and a normalization constraint, which not only can be used in most of algorithms directly without any extra modification but also reduces the memory requirement and computing cost significantly.

\begin{figure*}[tb]
\label{fig3}
\makebox[16.65cm][c]{
\begin{minipage}[t]{0.08\textwidth}
\centering
\includegraphics[width=0.96in]{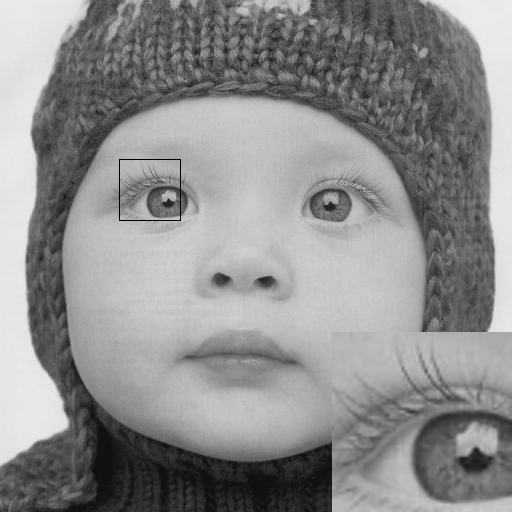}
\centering
\centering
\vskip -0.12 cm \begin{tiny}Original$\backslash$PSNR$\backslash$SSIM\end{tiny}
\end{minipage}
\hfill
\begin{minipage}[t]{0.08\textwidth}
\centering
\includegraphics[width=0.96in]{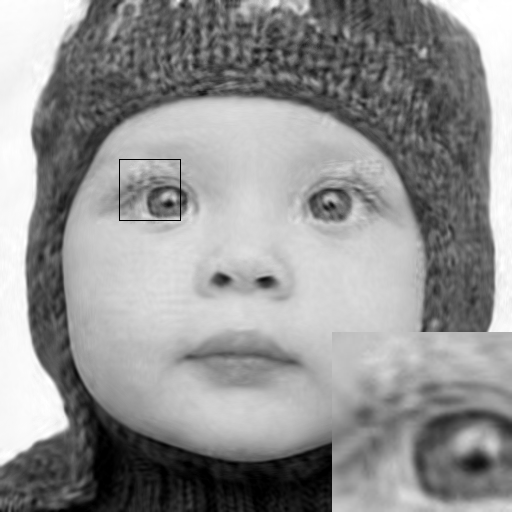}
\begin{scriptsize}
\centering
\vskip -0.4 cm \begin{tiny}GRM$\backslash$30.35$\backslash$0.9128\end{tiny}
\end{scriptsize}
\end{minipage}
\hfill
\begin{minipage}[t]{0.08\textwidth}
\centering
\includegraphics[width=0.96in]{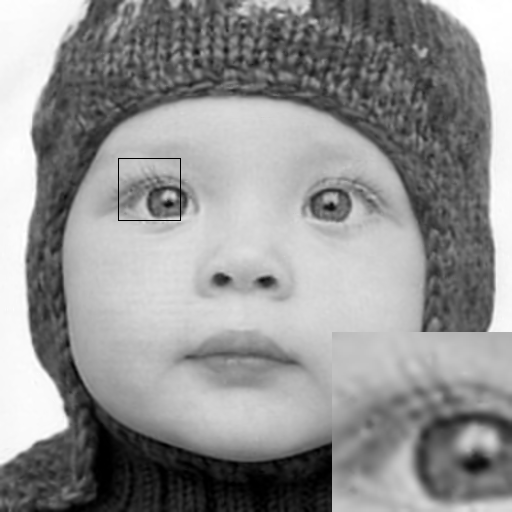}
\begin{scriptsize}
\centering
\vskip -0.4 cm \begin{tiny}DSMM(0.01)$\backslash$32.61$\backslash$0.9604\end{tiny}
\end{scriptsize}
\end{minipage}
\hfill
\begin{minipage}[t]{0.08\textwidth}
\centering
\includegraphics[width=0.96in]{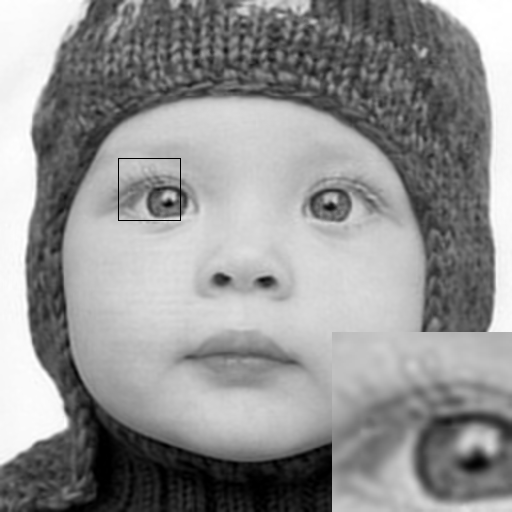}
\begin{scriptsize}
\centering
\vskip -0.4 cm \begin{tiny}DSMM(0.05)$\backslash$33.20$\backslash$0.9654\end{tiny}
\end{scriptsize}
\end{minipage}
\hfill
\begin{minipage}[t]{0.08\textwidth}
\centering
\includegraphics[width=0.96in]{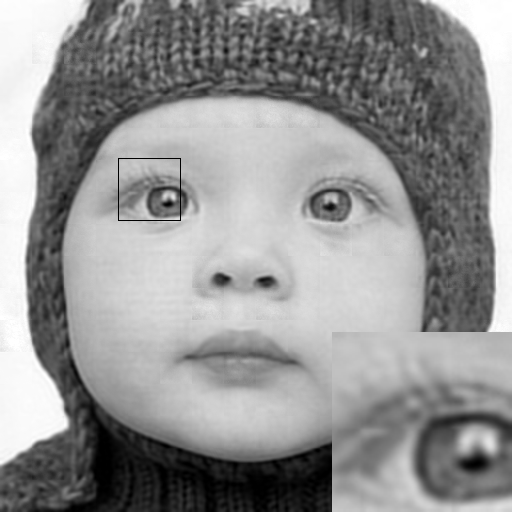}
\begin{scriptsize}
\centering
\vskip -0.4 cm \begin{tiny}DSMM(0.1)$\backslash$33.23$\backslash$0.9656\end{tiny}
\end{scriptsize}
\end{minipage}
\hfill
\begin{minipage}[t]{0.08\textwidth}
\centering
\includegraphics[width=0.96in]{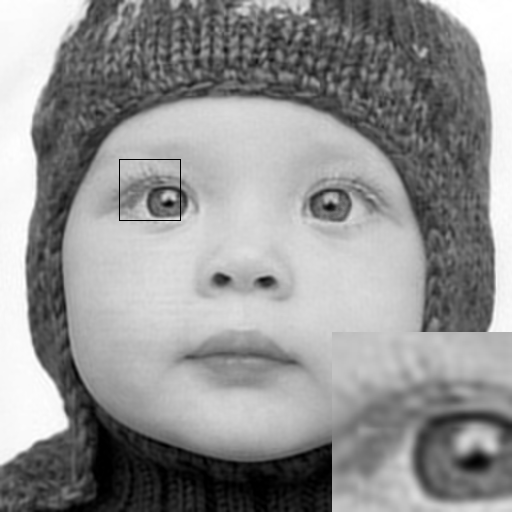}
\begin{scriptsize}
\centering
\vskip -0.4 cm \begin{tiny}DSMM(0.5)$\backslash$33.59$\backslash$0.9702\end{tiny}
\end{scriptsize}
\end{minipage}
\hfill
\begin{minipage}[t]{0.08\textwidth}
\centering
\includegraphics[width=0.96in]{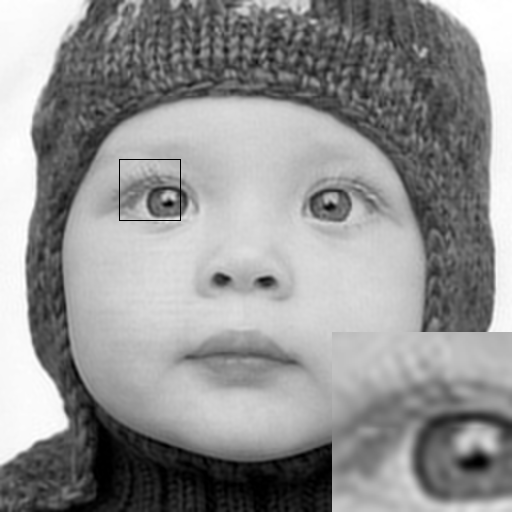}
\begin{scriptsize}
\centering
\vskip -0.40 cm \begin{tiny}DSMM(0.9)$\backslash$\textbf{33.61}$\backslash$\textbf{0.9707}\end{tiny}
\end{scriptsize}
\label{fig3}
\end{minipage}
}
\vspace{-1.41em}  \caption{{Visual quality comparison of Method MH~\cite{chen2011compressed} in the case of sampling ratio = 0.1 for image $\emph{Baby}$ in Set5}}
\label{Fig:fig3}
\end{figure*}

\subsubsection{Sparsity Constraint}

For $\boldsymbol{\Phi}$, we assume that the $k$th line of $\boldsymbol{\Phi}$ is denoted by $\Phi (k) = \{a_{k,1}, a_{k,2},\cdot\cdot\cdot, a_{k,N_{B}}\}$ and the sparsity degree by $\mathcal{SD}(\boldsymbol{\Phi}) = \frac{\mathcal{NUM}(\boldsymbol{\Phi},!0)}{N_{b}N_{B}}$, where $\mathcal{NUM}(\boldsymbol{\Phi},!0)$ indicates the numbers of nonzero elements in $\boldsymbol{\Phi}$ and $N_{b}N_{B}$ is the total elements in $\boldsymbol{\Phi}$. In order to produce the target sampling matrix with predefined sparsity degree $\mathcal{SD}(\boldsymbol{\Phi})=\alpha$ $(0 \leqslant \alpha < 1 )$, a sparsity constraint is defined as follows:

\begin{eqnarray}
\label{equ3}
\mathcal{S}(a_{k,i})=
\begin{cases}
0& \text{$|a_{k,i}|$ $\leqslant$ $\mu$}\\
a_{k,i}& \text{$|a_{k,i}|$ $>$ $\mu$}
\end{cases}
\end{eqnarray}
where $k = 1,2,\cdot\cdot\cdot,N_{b}$, $i = 1,2,\cdot\cdot\cdot,N_{B}$ and $\mu$ is the $(1-\alpha) N_{b} N_{B}$-$th$ smallest element in $|\boldsymbol{\Phi}|$. Fig.~\ref{Fig:fig2} shows the details of the sparsity constraint and its approximate derivative. Through this constraint, expected $\boldsymbol{\Phi}$ with different sparsity degree can be generated, which reduces the storage cost and computational complexity greatly.

\subsubsection{Normalization Constraint}

In most of CS literatures~\cite{li1tval3,chen2011compressed,zhang2012compressed,zhang2014group}, the measurement matrix is normalized to control the range of measurements. Mathematically, the normalization constraint of $\boldsymbol{\Phi}$ is defined by:
\begin{eqnarray}
\label{equ4}
  \sum_{i=1}^{N_{B}}a_{k,i}^{2} = 1, \ \ k = 1,2,\cdot\cdot\cdot,N_{b}
\end{eqnarray}
In fact, $\Phi (k)$ is the $k$th kernel of convolutional layer in the sampling sub-network and $a_{k,i}$ is the $i$th value of this kernel. In this paper, in order to generate normalized sampling matrix, we applied a normalization constraint to the parameters of the convolutional layer in the sampling sub-network. The normalization constraint for $k$th kernel can be formulated as:
\begin{eqnarray}
\label{equ5}
  \mathcal{F}(s_{k,j}) = \frac{s_{k,j}}{\sqrt{\sum_{i=1}^{N_{B}}s_{k,i}^{2}}},\ \ j = 1,2,\cdot\cdot\cdot,N_{B}
\end{eqnarray}
where $s_{k,j}=\mathcal{S}(a_{k,j})$ and its derivative executed as:
\begin{eqnarray}
\label{equ6}
  \mathcal{F}^{'}(s_{k,j}) = \frac{\sqrt{\omega}-\frac{s_{k,j}^{2}}{\sqrt{\omega}}}{\sqrt{\omega}},\ \ \ \omega = \sum_{i=1}^{N_{B}}s_{k,i}^{2}
\end{eqnarray}
Through Eq.~\ref{equ5}, we got the normalized parameters, namely the normalized sampling matrix. Then this sampling matrix will be used to sample the original images.

%
%
%
%
%
%
%
%

\begin{table}[!t]

\caption{Quantitative evaluation of GRM and DSMM for method GSR: Average PSNR$\backslash$SSIM at different sampling ratios 0.1, 0.2 and 0.3 on datasets Set5. {\color{red}Red} text indicates the best performance}

\centering
\begin{tabular}{>{\hfil}p{58pt}<{\hfil}|>{\hfil}p{45pt}<{\hfil}|>{\hfil}p{45pt}<{\hfil}|
    >{\hfil}p{45pt}<{\hfil}}
\hline
\hline
Alg. & \multicolumn{3}{c}{GSR~\cite{zhang2014group}}\\
\cline{1-4}
sampling ratio    &   0.1  & 0.2 & 0.3\\
\hline
\hline
GRM    &   30.6$\backslash$0.88  & 34.4$\backslash$0.92 & 37.1$\backslash$0.95\\

DSMM-1\%    &  32.0$\backslash$0.88  & 34.9$\backslash$0.93 & 37.1$\backslash$0.95\\

DSMM-2\%    &  32.6$\backslash$0.89  & 35.5$\backslash$0.93 & 37.8$\backslash$0.96\\

DSMM-5\%    &  32.7$\backslash$0.90  & 35.6$\backslash$0.94 & 38.1$\backslash$0.96\\


DSMM-20\%    &  $\textbf{\color{red}33.0}\backslash\textbf{\color{red}0.91}$  & $\textbf{\color{red}36.1}\backslash\textbf{\color{red}0.95}$ & $\textbf{\color{red}38.5}\backslash\textbf{\color{red}0.97}$\\

DSMM-50\%    &  32.9$\backslash\textbf{\color{red}0.91}$  & $\textbf{\color{red}36.1}\backslash\textbf{\color{red}0.95}$ & 38.4$\backslash\textbf{\color{red}0.97}$\\


DSMM-90\%    &  32.9$\backslash\textbf{\color{red}0.91}$  & 36.0$\backslash\textbf{\color{red}0.95}$ & 38.4$\backslash\textbf{\color{red}0.97}$\\

DSMM    &   32.9$\backslash\textbf{\color{red}0.91}$  & 36.0$\backslash\textbf{\color{red}0.95}$ & 38.4$\backslash\textbf{\color{red}0.97}$\\

\hline
\textbf{Gain}    & $\textbf{+2.4}\backslash\textbf{0.03}$  & $\textbf{+1.7}\backslash\textbf{0.03}$ & $\textbf{+1.4}\backslash\textbf{0.02}$\\

\hline
\hline

\end{tabular}
\label{Tab:tab2}
\end{table}

\begin{table*}[!t]

\caption{Quantitative evaluation of GRM and DSMM for method MH and CoS: Average PSNR$\backslash$SSIM at different sampling ratios 0.1, 0.2 and 0.3 on datasets Set5. {\color{red}Red} text indicates the best performance}

\centering
\begin{tabular}{>{\hfil}p{58pt}<{\hfil}|>{\hfil}p{48pt}<{\hfil}|>{\hfil}p{48pt}<{\hfil}|
    >{\hfil}p{48pt}<{\hfil}|>{\hfil}p{48pt}<{\hfil}|>{\hfil}p{48pt}<{\hfil}|>{\hfil}p{48pt}<{\hfil}|>{\hfil}p{48pt}<{\hfil}}
\hline
\hline
Alg. & \multicolumn{3}{c|}{MH~\cite{chen2011compressed}} &  \multicolumn{3}{c|}{CoS~\cite{zhang2012compressed}} & \multirow{2}*{Avg.}\\
\cline{1-7}
sampling ratio    &   0.1  & 0.2 & 0.3 & 0.1 & 0.2 & 0.3\\
\hline
\hline
GRM    &   27.4$\backslash$0.803  & 30.7$\backslash$0.881 & 32.6$\backslash$0.911 & 28.3$\backslash$0.819  & 30.0$\backslash$0.858 & 30.6$\backslash$0.871 & 29.9$\backslash$0.857\\

DSMM-1\%    &   29.0$\backslash$0.853  & 31.9$\backslash$0.905 & 32.3$\backslash$0.911 & 30.1$\backslash$0.864  & 32.2$\backslash$0.898 & 33.5$\backslash$0.910 & 31.5$\backslash$0.890\\

DSMM-2\%    &   29.3$\backslash$0.861  & 32.3$\backslash$0.913 & 32.5$\backslash$0.919 & 30.4$\backslash$0.872  & 32.7$\backslash$0.905 & 33.8$\backslash$0.920 & 31.8$\backslash$0.898\\

DSMM-5\%    &   29.7$\backslash$0.868  & 32.7$\backslash$0.921 & 32.9$\backslash$0.922 & 30.7$\backslash$0.878  & 33.0$\backslash$0.914 & 34.4$\backslash$0.927 & 32.2$\backslash$0.905\\


DSMM-20\%    &   30.0$\backslash$0.873  & 33.0$\backslash$0.925 & 33.2$\backslash$0.930 & 31.0$\backslash$0.884  & 33.4$\backslash$0.918 & 34.6$\backslash$0.930 & 32.5$\backslash$0.910\\

DSMM-50\%    &   $\textbf{\color{red}30.1}\backslash$0.880  & $\textbf{\color{red}33.3}\backslash\textbf{\color{red}0.934}$ & $\textbf{\color{red}33.5}\backslash\textbf{\color{red}0.932}$ & $\textbf{\color{red}31.1}\backslash\textbf{\color{red}0.888}$  & $\textbf{\color{red}33.7}\backslash\textbf{\color{red}0.926}$ & $\textbf{\color{red}34.8}\backslash\textbf{\color{red}0.932}$ & $\textbf{\color{red}32.8}\backslash\textbf{\color{red}0.915}$\\


DSMM-90\%    &   $\textbf{\color{red}30.1}\backslash\textbf{\color{red}0.881}$  & $\textbf{\color{red}33.3}\backslash\textbf{\color{red}0.934}$ & 33.3$\backslash$0.930 & 31.0$\backslash$0.886  & 33.5$\backslash$0.923 & 34.6$\backslash$0.930 & 32.6$\backslash$0.914\\

DSMM    &   $\textbf{\color{red}30.1}\backslash\textbf{\color{red}0.881}$  & 33.1$\backslash\textbf{\color{red}0.934}$ & 33.3$\backslash$0.931 & 30.9$\backslash$0.884  & 33.4$\backslash$0.921 & 34.7$\backslash$0.931 & 32.6$\backslash$0.914\\

\hline
\textbf{Gain}    &   $\textbf{+2.7}\backslash\textbf{0.078}$  & $\textbf{+2.6}\backslash\textbf{0.054}$ & $\textbf{+0.9}\backslash\textbf{0.021}$ & $\textbf{+2.8}\backslash\textbf{0.069}$ & $\textbf{+3.7}\backslash\textbf{0.068}$ & $\textbf{+4.2}\backslash\textbf{0.061}$ & $\textbf{+2.9}\backslash\textbf{0.058}$\\

\hline
\hline

\end{tabular}
\label{Tab:tab1}
\end{table*}

\subsection{Reconstruction Sub-network}

The measurements is hardly to be used to evaluate the image reconstruction quality directly. In order to improve the CS reconstruction performance, the image reconstruction sub-network are utilized to help enhance the sampling sub-network. For the reconstruction algorithms of CS, many optimization-based~\cite{li1tval3,chen2011compressed,zhang2012compressed,zhang2014group} and deep learning based methods~\cite{mousavi2015deep,kulkarni2016reconnet,shi2017deep} have been proposed. Considering the hardness of calculating derivatives for the optimization-based methods and inspired by the aforementioned deep learning based works, we utilize a ``reshape+concat'' layer and several convolutional layers to reconstruct original images as shown in Fig.~\ref{Fig:fig1}.
Given the compressed measurement vector, we utilize a ``reshape+concat'' layer~\cite{shi2017deep} to obtain a feature map with the same size as the original block.
Then the further nonlinear reconstruction is executed by using several convolutional layers, which has the same configuration with CSNet~\cite{shi2017deep} (d=3). In our network, this reconstruction sub-network has two important missions. One is that reconstructing the target images from its measurements. Another is providing guidance for the optimization of sampling sub-network.

\section{Training Details and Experiments}
\label{sec:pagestyle}

In our model, the sampling sub-network and the reconstruction sub-network are optimized jointly. In this section, we describe the training details as well as the experimental results. Given the input image block $x$, our goal is to obtain highly compressed measurement $y$ with the sampling sub-network, and then accurately recover it to the original input image block $x$ with the reconstruction sub-network. The input and the label are all image block $x$ itself for training. Therefore, the training dataset can be represented as $\{x_{i},x_{i}\}_{i}^{N}$. The mean square error (MSE) is adopted as the cost function of our network. The optimization objective is represented as
\begin{eqnarray}
\label{equ8}
 \min \frac{1}{2N}\sum_{i=1}^{N}\|g(x_{i};\theta_{s},\theta_{r})-x_{i}\|_{2}^{2}
\end{eqnarray}
where $g$ is the operations of our network. $\theta_{s}$ and $\theta_{r}$ is the parameters of sampling sub-network and reconstruction sub-network. $g(x_{i};\theta_{s},\theta_{r} )$ is the final CS reconstructed output with respect to $x_{i}$.

    \begin{algorithm}[h]
        \caption{The training process of the proposed network for one iteration}
        \begin{algorithmic}[0] 
            \Ensure Initiate the parameters $\theta_{s}$, $\theta_{r}$ of network
            \Require $\{ x_{i}$, $x_{i} \}$\\
            $\theta_{G} \gets global\ variable$
            \Function {Forward}{$x_{i}$}
                \State $\theta_{G} \gets \theta_{s}$
                \State $\theta_{s} \gets sparsity\ constraint$
                \For{$kernel$ in $\theta_{s}$}
                    \State $kernel \gets normalization\ constraint$
                \EndFor
                \State \Return{$Output \gets \|g(x_{i},\ \theta_{s},\ \theta_{r})-x_{i}\|_{2}^{2}$}
            \EndFunction
            \Function{Backward}{$Output,\ x_{i}$}
                \State $\frac{d(Output)}{d(\theta_{s})},\ \frac{d(Output)}{d(\theta_{r})} \gets \nabla(x_{i},\ Output)$
                \State $\theta_{r} \gets \theta_{r}-\gamma \frac{d(Output)}{d(\theta_{r})}\ \ \ (\gamma \ is\ learning\ rate)$
                \State $\theta_{s} \gets \theta_{G}$
                \State $\theta_{s} \gets \theta_{s}-\gamma (\frac{d(Output)}{d(\theta_{s})}*\mathcal{F}^{'}(\mathcal{S}(\theta_{s})))$
            \EndFunction
            \renewcommand{\algorithmicrequire}{\textbf{Note:}}
            \Require $\nabla\ is\ the\ gradient\ back\ propagation.$
        \end{algorithmic}
    \end{algorithm}

We use the training set of the BSDS500 database~\cite{arbelaez2011contour} for training, and the its validation set for validation. We set the patch size as 96$\times$96, batch size as $32$ and block size as 32 ($B$=32). We augment the training data in two ways: $(i)$ Randomly scale between $[0.8, 1.2]$. $(ii)$ Flip the images horizontally with a probability of 0.5. The DSMM is trained with the Matlab toolbox MatConvNet~\cite{vedaldi2015matconvnet} on a Titan X GPU. The momentum parameter is set as 0.9 and weight decay as $1e$-$4$. We train our model for 100 epochs and each epoch iterates 600 times. Standard gradient descent (SGD) is used to optimize all network parameters. For the learning rate, the first 30 epochs is set $1e$-$3$. The following 40 epochs are declined equably from $1e$-$4$ to $1e$-$6$ and the last 30 epochs is set as $1e$-$6$. Algorithm 1 shows the details of training process of our framework for one iteration.

We evaluate the performance of the proposed DSMM for CS reconstruction in the algorithms MH~\cite{chen2011compressed}, CoS~\cite{zhang2012compressed} and GSR~\cite{zhang2014group}. Specifically, we replace the gaussian random matrix with our DSMM and recover the original images with these methods. To evaluate the performance of each algorithm, we investigate three different sampling ratios 0.1, 0.2, 0.3 and seven different sparse degrees 0.01(DSMM-1\%), 0.02(DSMM-2\%), 0.05(DSMM-5\%), 0.2(DSMM-20\%), 0.5(DSMM-50\%), 0.9(DSMM-90\%), 1.0 (DSMM) with assessment criteria PSNR and SSIM. The comparisons with various algorithms on benchmark Set5 are provided in Table~\ref{Tab:tab2} and Table~\ref{Tab:tab1}. Some visual comparisons shown in Figs.~\ref{Fig:fig3}. demonstrate that our proposed DSMM preserves much sharper edges and finer details. In regular CS applications, DSMM-20\% has almost achieved the best performances.

\section{Conclusion}
\label{sec:majhead}

In this paper, a novel deep neural network based sparse measurement matrix (DSMM) for CS sampling of natural images is proposed by utilizing a deep convolutional network, which not only ensure sufficient information in the measurements for CS reconstruction but also reduces the memory requirement and computing cost significantly. Experimental results show that the proposed DSMM greatly enhances the existing performance of CS reconstruction when compared with GRM.

\section{Acknowledgements}

This work is partially funded by the Major State Basic Research Development Program of China (973 Program 2015CB351804) and the National Natural Science Foundation of China under Grant No. 61572155 and 61672188.
%


\bibliographystyle{IEEEtran}
\bibliography{refs}

\end{document}